\documentclass[sigconf]{acmart}
\usepackage{url}
\usepackage{float}
\AtBeginDocument{%
  }

\copyrightyear{2025}
\acmYear{2025}
\setcopyright{cc}
\acmConference[UMAP Adjunct '25]{Adjunct Proceedings of the 33rd ACM
Conference on User Modeling, Adaptation and Personalization}{June 16--19,
2025}{New York City, NY, USA}
\acmBooktitle{Adjunct Proceedings of the 33rd ACM Conference on User
Modeling, Adaptation and Personalization (UMAP Adjunct '25), June 16--19,
2025, New York City, NY, USA}
\acmDOI{10.1145/3708319.3734174}
\acmISBN{979-8-4007-1399-6/2025/06}

\usepackage{makecell}

\begin{document}

\title[Differentiable Fuzzy Neural Networks for Recommender Systems]{Differentiable Fuzzy Neural Networks for Recommender Systems}

\author{Stephan Bartl}
\email{stephan.bartl@student.tugraz.at}
\affiliation{%
  \institution{Graz University of Technology}
  \city{Graz}
  \country{Austria}
}
\author{Kevin Innerebner}
\email{innerebner@tugraz.at}
\affiliation{%
  \institution{Graz University of Technology}
  \city{Graz}
  \country{Austria}
}
\author{Elisabeth Lex}
\email{elisabeth.lex@tugraz.at}
\affiliation{%
  \institution{Graz University of Technology}
  \city{Graz}
  \country{Austria}
}


\begin{abstract}
As recommender systems become increasingly complex, transparency is essential to increase user trust, accountability, and regulatory compliance.
Neuro-symbolic approaches that integrate symbolic reasoning with sub-symbolic learning offer a promising approach toward transparent and user-centric systems.
In this work-in-progress, we investigate using fuzzy neural networks (FNNs) as a neuro-symbolic approach for recommendations that learn logic-based rules over predefined, human-readable atoms. Each rule corresponds to a fuzzy logic expression, making the recommender's decision process inherently transparent. In contrast to black-box machine learning methods, our approach reveals the reasoning behind a recommendation while maintaining competitive performance. 
We evaluate our method on a synthetic and MovieLens 1M datasets and compare it to state-of-the-art recommendation algorithms.
Our results demonstrate that our approach accurately captures user behavior while providing a transparent decision-making process.
Finally, the differentiable nature of this approach facilitates an integration with other neural models, enabling the development of hybrid, transparent recommender systems.
\footnote{Code to reproduce the experiments available at \url{https://github.com/stephba/diff-fnn}}
\end{abstract}




\maketitle

\section{Introduction}
In the digital age, recommender systems play a crucial role in helping users navigate vast information spaces, enhancing user experience across domains such as e-commerce, streaming services, and social media. 
While state-of-the-art approaches based on deep learning~\cite{NeuroCollabFilt2017, 10.1145/3397271.3401063} or matrix factorization~\cite{rendle2012bprbayesianpersonalizedranking} achieve strong predictive performance, they often lack transparency and operate as opaque black boxes~\cite{10.1145/3631700.3658522}.
Such a lack of transparency can limit user trust and accountability and complicate efforts to align system behavior with ethical or regulatory requirements~\cite{schedl2025technical}. 

Researchers have started tackling this problem in recent years by developing neuro-symbolic approaches that combine neural networks with symbolic reasoning. 
For instance, \citet{RecGraphEmb2024} enrich graph embeddings of knowledge graphs by incorporating symbolic rule mining.
Another neuro-symbolic approach combines propositional expressions from user history with graph neural networks \cite{NeuroGraph2023}.
Additionally, \citet{NeuroCollab2021} suggest encoding user-item interactions using neural networks and feeding these encodings into a symbolic reasoning part where boolean operations are implemented as neural networks and truth values are high-dimensional vectors.
Similarly, \citet{10.1007/978-3-031-27181-6_8} use high-dimensional tensors to model logical reasoning to inject background knowledge to deal with data sparsity and new users without an interaction history.
Although these approaches introduce symbolic components, they rely on deep neural architectures or latent representations, making their decision processes difficult to interpret from a human perspective. 

In this paper, we propose a transparent neuro-symbolic recommendation model based on fuzzy neural networks (FNNs).
FNNs combine the continuous reasoning capabilities of fuzzy logic~\cite{zadeh1975fuzzy} with the learning capacity of neural network architectures. Our approach generates recommendations by learning weighted logic rules over a set of predefined, human-readable propositional atoms. Each rule is implemented as a fuzzy logic expression over the continuous range of $[0,1]$, using differentiable fuzzy operators (e.g., AND, OR)~\cite{lin1996neural} defined via triangular norms (i.e., t-norms)~\cite{klement2013triangular}. These differentiable operators support smooth training with gradient descent while preserving a clear logical interpretation.



Unlike previous research on FNNs for recommender systems~\cite{ochoa2021medical, zhang2021causal, sinha2020building, deng2024novel, rutkowski2018content}, our model is built entirely from differentiable components that model fuzzy logic operations like AND and OR over the continuous range $[0,1]$, i.e., \textit{fuzzy neurons}~\cite{lin1996neural}, eliminating the need for additional methods or algorithms. These neurons are used to learn logic rules over a set of predefined atoms. While the fuzzy weights are optimized during training, the atoms remain fixed, ensuring that the learned rules remain transparent. 

We demonstrate the feasibility of our approach on a synthetic dataset, where the ground truth rules are known, and the MovieLens 1M dataset. Our results show that our model achieves competitive performance against standard baselines, suggesting that fuzzy neural networks are a promising foundation for building human-centered, transparent recommender systems.

\section{Methodology}
Here, we outline the fuzzy operators, the model architecture, the datasets used, the evaluation process, and the experimental setup.

\subsection{Fuzzy Operators}

Our model relies on fuzzy neurons to implement logical operations. These neurons represent fuzzy operators such as NOT, AND, and OR over the continuous range $[0,1]$~\cite{lin1996neural}.
The operators can be defined via triangular norms (i.e., t-norms) and their dual conorms, which operate on continuous values in $[0,1]$. Popular examples are the minimum t-norm $T_M(a,b)=min(a,b)$, the Łukasiewicz t-norm $T_L(a,b)=max(a+b-1,0)$, or the product t-norm $T_P(a,b)=a\cdot b$~\cite{klement2013triangular}. 
In our work, we leverage the differentiable and smooth product t-norm for the fuzzy operators to enable the use of gradient-based optimization mechanisms. While other t-norms, such as Łukasiewicz or minimum t-norm, could also be used, they may introduce non-differentiable points or vanishing gradient issues.

The fuzzy operators according to the product t-norm are defined as follows for $a,b \in [0,1]$:
\begin{itemize}
    \item $NOT(a) = 1 - a$
    \item $AND(a,b) = a \cdot b$
    \item $OR(a,b) = a + b - a \cdot b$
\end{itemize}

For compact notation, we also define vectorized versions of operations: 

$\mathbf{a},\mathbf{b} \in [0,1]^n$:
\begin{itemize}
    \item $AND(\mathbf{a}, \mathbf{b}) = \mathbf{a} \odot \mathbf{b}$ (Hadamard product)
    \item $AND(\mathbf{a}) = \Pi_{i=1}^n a_i$
    \item $OR(\mathbf{a},\mathbf{b}) = \mathbf{a} + \mathbf{b} - \mathbf{a} \odot \mathbf{b}$
    \item $OR(\mathbf{a}) = OR(OR(\cdots OR(OR(a_1,a_2), a_3),\cdots), a_n)$
\end{itemize}
Please note that $OR(\mathbf{a})$ corresponds to left-associative binary OR operations.

\subsection{Model Architecture}
Since our model only operates on values between $0$ and $1$, we formulate the recommendation task as a binary classification problem:
Given a user-item pair as input, the model should output $1$ if the item is relevant to the user and $0$ if it is not.

\begin{figure}[h]
  \centering
  \includegraphics[width=\linewidth,alt={
  This figure depicts our model architecture that processes user and item inputs.
  First, predefined atoms are generated, which are then fed into several rule layers, implementing the weighted conjunctions.
  Finally, the layer outputs are aggregated via an OR neuron to produce the model's final output.
  }]{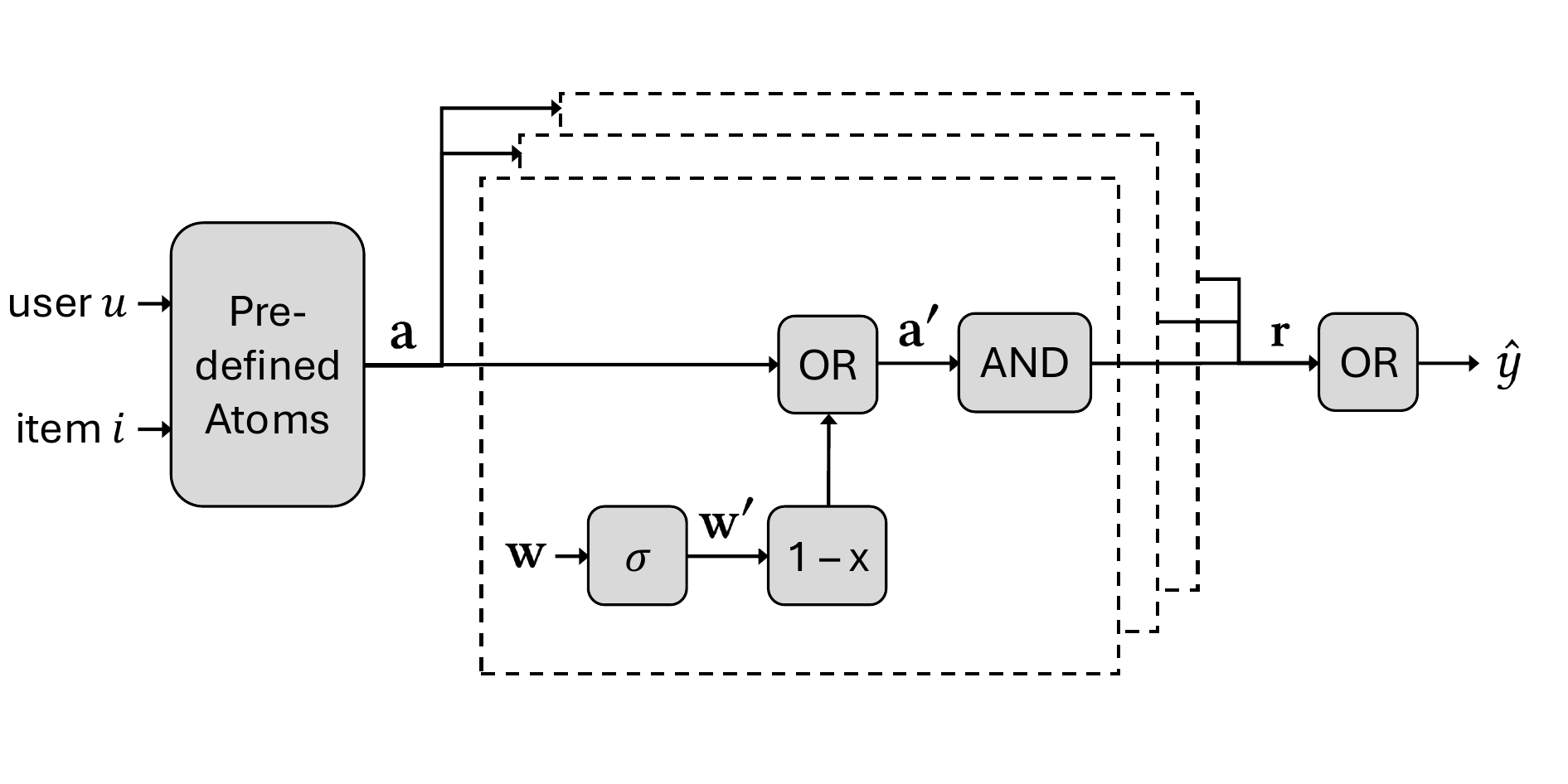}
  \caption{Model architecture: Given a user-item pair as input, the model first computes predefined propositional atoms. These atoms are then fed into parallel layers, implementing fuzzy decision rules based on fuzzy neurons. The outputs of the layers are then aggregated via a fuzzy OR neuron, leading to the model's final output. This output classifies the item as either relevant or not relevant to the user.}
  \label{fig:arch}
\end{figure}

Figure~\ref{fig:arch} illustrates the architecture of our model. The input to the model consists of fuzzy propositional atoms $\mathbf{a}\in [0,1]^n$. Each atom represents a human-readable feature that captures a specific property of the user, the item, or their interaction. For example, whether an item has a high average rating, whether the item's genre matches the preferences of the user, or whether the user has rated similar items in the past.  

These atoms are computed using fixed, predefined functions and do not change during training. That is, they are not learned from data but instead deterministically derived from available metadata and interaction statistics. The exact definitions of the atoms used in our experiments are described in Section~\ref{sec:synthetic_dataset} for the synthetic dataset and Section~\ref{sec:movielens_dataset} for the MovieLens 1M dataset.

The atoms $\mathbf{a}$ are then fed into $k$ parallel layers, each representing a different fuzzy decision rule. This models the idea that a user might consider multiple independent reasoning paths when deciding whether an item is relevant.
Each rule is implemented using fuzzy logic operators, i.e.,  AND and OR neurons, and computes a weighted conjunction of the atoms using learned fuzzy weights $\mathbf{w'}_i\in [0,1]^n$ for $i \in \{1, \ldots, k\}$.
The weights reflect each atom's contribution to a rule and can be compactly written as a fuzzy weight matrix:
\[
\mathbf{W'} = \begin{bmatrix}
\mathbf{w'}_1 & \mathbf{w'}_2 & \dots & \mathbf{w'}_k
\end{bmatrix}^T \in [0,1]^{k \times n}.
\]
To compute each rule, atoms $\mathbf{a}$ and fuzzy weights $\mathbf{w'}$ are combined using an elementwise OR operator as proposed by~\citet{pedrycz1993fuzzy}, resulting in a vector of weighted atoms $\mathbf{a'}_i= OR(\mathbf{a}, \mathbf{1} - \mathbf{w'}_i) \in [0,1]^n$.
This ensures that atoms with a higher weight contribute more strongly to the rule, i.e., via  $\mathbf{1} - \mathbf{w'}_i$.
The weighted atoms are then aggregated using a fuzzy AND operator, resulting in the rule output $\mathbf{r_i} = AND(\mathbf{a'}_i)\in [0,1]$.
Finally, the outputs of all $k$ rules are combined using a fuzzy OR operator to produce the model's prediction $\hat{y} = OR(\mathbf{r})\in [0,1]$.
Instead of optimizing the fuzzy weights directly, we train real-valued weights $\mathbf{w}_i \in \mathbb{R}^n$ using gradient descent.
These real-valued weights are then mapped to fuzzy weights by applying the sigmoid function: $\mathbf{w'}_i = \sigma(\mathbf{w}_i)$.

Equation \ref{eq:bool} presents the propositional formula $F$ modeled by this network architecture:
\begin{equation}
\label{eq:bool}
F = \bigvee_{i=1}^k R_i \quad\text{with } R_i = \bigwedge_{j=1}^n (a_j \lor (1 - \mathbf{W'}_{ij}))
\end{equation}
Under the closed-world assumption~\cite{REITER1981119}, this corresponds to a set of horn clauses~\cite{horn1951sentences} of the form $(a'_1\land a'_2 \land \dots)\to RELEVANT$. This means that unless one of the learned rules suggests an item is relevant, the atom $RELEVANT$ is considered $False$. In other words, the set of learned decision rules is assumed to be complete.

To train the model, we generate binary targets by applying a rating threshold: user-item pairs with high ratings are labeled as relevant (i.e., $y=1$), and those with low ratings are labeled as not relevant (i.e.,~$y=0$). 
As training loss, we compute the mean squared error (MSE) between the predicted labels $\mathbf{\hat{y}}\in [0,1]^N$ and the target values $\mathbf{y}\in [0,1]^N$ over $N$ training samples.
Additionally, we apply L1 regularization on the fuzzy weight matrix $\mathbf{W'}$ with regularization parameter $\lambda$. Since the regularization term grows linearly with the number of atoms $n$ and the number of rules $k$, we introduce a normalization by $\frac{1}{kn}$.
The final objective function is given in Equation \ref{eq:objective}.
\begin{equation}
\label{eq:objective}
\mathcal{L}(\mathbf{w}) = \frac{1}{N} \sum_{i=1}^{N} (\hat{y}^{(i)} - y^{(i)})^2 + \lambda\frac{1}{kn}\|\mathbf{W'}\|_1
\end{equation}

\subsection{Datasets}
We investigate our approach using a synthetic dataset to illustrate the concept and verify the ability to learn correct rules.
Additionally, we test its performance on the MovieLens 1M dataset~\cite{harper2015movielens}.
We sort the MovieLens 1M dataset by temporal order and then split it according to a ratio, which leads to unseen users and items, i.e., cold starts, in the test data.
The synthetic dataset is divided using a random split. 
Both datasets are partitioned into training, validation, and test sets with a ratio of $0.7$/$0.1$/$0.2$. 
More information about the datasets used can be found in Table~\ref{tab:dataset}.

\begin{table}
\caption{Dataset statistics}
\label{tab:dataset}
\begin{tabular}{lllll}
\toprule
Dataset & \#Users & \#Items & \#Interactions & Density \\
\midrule
Synthetic & 6,040 & 3,883 & 1,000,209 & 4.26\% \\
MovieLens 1M & 6,040 & 3,883 & 1,000,209 & 4.26\%\\
\bottomrule
\end{tabular}
\end{table}

\subsubsection{Synthetic Dataset}
\label{sec:synthetic_dataset}
The synthetic dataset is used to verify if the model can reproduce ground truth rules.
Our toy example consists of 6,040 simulated users and 3,883 simulated items with 1 million ratings and is, hence, the same size as the MovieLens 1M dataset.
We frame this example as a movie recommendation task with the following artificial propositional atoms:
\begin{itemize}
    \item $\text{HIGH}$: The movie has a high rating
    \item $\text{GENRE}$: The genre of the movie matches the user's favorite genre
    \item $\text{RECENT}$: The movie is recent
    \item $\text{CAST}$: The movie features the favorite actress or actor of the user
    \item $\text{DIRECTOR}$: The movie is directed by the user's favorite director
    \item $\text{COOKIE}$: The user likes cookies (random atom)
    \item $\text{RELEVANT}$: The movie is relevant to the user
\end{itemize}
To balance positive and negative samples, the probability that an item is relevant given a user-item pair is $0.5$.
The data is generated such that each rule has an equal number of user-item pairs supporting it.
The target values for the training, validation, and test set are then generated using the following propositional formula:
\begin{align*}
RELEVANT \equiv \; & HIGH\\
&\lor (RECENT \land GENRE) \\
&\lor (RECENT \land CAST \land DIRECTOR)
\end{align*}
Under the closed-world assumption, this is equivalent to these horn clauses:
\begin{align*}
HIGH \to RELEVANT \\
RECENT \land GENRE \to RELEVANT \\
RECENT \land CAST \land DIRECTOR \to RELEVANT
\end{align*}
\subsubsection{MovieLens 1M}
\label{sec:movielens_dataset}
The MovieLens 1M dataset is a widely used benchmark in recommender system research. It consists of approximately 1 million movie ratings with a range between $1$ and $5$ rated by 6,040 users across 3,883 movies. 
To prepare the dataset for training our model, we generate positive and negative samples based on a rating threshold of $4.0$, i.e., assigning ratings below $4.0$ with a target value of $0$ and ratings greater than or equal to $4.0$ with a target value of $1$.

\subsection{Evaluation}
The model's performance is evaluated using the following metrics: Precision@k, Recall@k, NDCG@k (Normalized Discounted Cumulative Gain), and MAP@k (Mean Average Precision), each calculated for $k=5$ and $k=10$. Please note that while our model outputs binary labels, we treat the output probabilities as confidence scores to rank items per user. We use these rankings to compute our performance metrics in a top-k evaluation setup.

Our approach is compared against the following baseline models to evaluate its performance:

\begin{itemize}
    \item \textbf{BaselineOnly}~\cite{Hug2020}: Calculates ratings for a given user and item by considering the global average rating, user bias, and item bias: $\hat{r}_{ui} = b_{ui} = \mu + b_u + b_i$.
    \item \textbf{KNNBasic}~\cite{Hug2020}: Determines the average rating of the $k=40$ nearest neighbors, leveraging collaborative filtering to make predictions.
    \item \textbf{SVD}~\cite{Hug2020}: A matrix factorization algorithm that applies Singular Value Decomposition (SVD) to the user-item interaction matrix.
    \item \textbf{LightGCN}~\cite{10.1145/3397271.3401063}: A recommendation model employing a graph convolutional network for collaborative filtering.
    \item \textbf{JRip}~\cite{frank2016weka}: Java implementation of the symbolic propositional rule learning algorithm \textit{Repeated Incremental Pruning to Produce Error Reduction (RIPPER)} as proposed by \citet{Cohen1995}. Input for this baseline are the predefined propositional atoms.
    \item \textbf{Decision Tree}: As decision trees can handle both propositional as well as numerical and categorical input, this baseline is evaluated twice: first using the original item and user features from the dataset (e.g., gender, age, movie genre), and then using the predefined atoms as input.
    \item \textbf{NCF}~\cite{NeuroCollabFilt2017}: Neural Collaborative Filtering (NCF) is a state-of-the-art recommendation approach that joins matrix factorization with neural networks, which can capture complex patterns in user-item interactions. The hyperparameter settings are taken from the original paper by \citet{NeuroCollabFilt2017}.  
\end{itemize}

\subsection{Experimental Setup}
Our neural network is built with PyTorch and PyG~\cite{fey2019fast}, while baselines are implemented using the Surprise library~\cite{Hug2020}, the Waikato Environment for Knowledge Analysis (WEKA)~\cite{frank2016weka}, the Microsoft Recommenders library~\cite{10.1145/3298689.3346967} and the scikit-learn library.

Model parameters are optimized using Adam optimizer~\cite{kingma2015adam}.
As our model and some baselines include randomness due to weight initialization or sampling, the experiments are repeated 10 times to compute statistical significance.

Hyperparameter values are found through exploratory experiments and are selected based on the validation results, convergence of validation loss, and sparseness of the learned rules.
The tested values for the number of rules are $k \in \{2,4,8,16\}$, the learning rate is tested over a range between $0.001$ and $0.1$, the number of epochs over a range between $50$ and $1000$, and the regularization parameter $\lambda$ over the range $[0.01, 1.0]$.
However, comprehensive hyperparameter tuning still has to be conducted in future work to further investigate the robustness of this approach.

All experiments were conducted on a MacBook with an Apple M1 Pro chip and 32GB of RAM.

\section{Preliminary Results}
In this section, we present the learned rules and our results.

\subsection{Synthetic Dataset}

\begin{table}[htp]
\caption{Fuzzy weight matrix $\mathbf{W'}$ of the synthetic dataset for rules R1, R2, R3, R4. Values above 0.1 highlighted in bold.}
\label{tab:fuzzysynth}
\centering
\setlength{\tabcolsep}{4pt} 
\begin{tabular}{lcccccc}
\toprule
 & \makecell{RECENT} & \makecell{GENRE} & \makecell{CAST} & \makecell{DIRECTOR} & \makecell{HIGH} & \makecell{COOKIES} \\
\midrule
R1 & $\mathbf{0.994}$ & $\mathbf{0.987}$ & 0.013 & 0.004 & 0.008 & 0.003 \\
R2 & 0.005 & 0.009 & 0.019 & 0.008 & $\mathbf{0.993}$ & 0.013 \\
R3 & $\mathbf{0.986}$ & 0.002 & $\mathbf{0.927}$ & $\mathbf{0.915}$ & 0.002 & 0.005 \\
R4 & 0.029 & 0.003 & 0.020 & 0.003 & $\mathbf{0.990}$ & 0.004 \\
\bottomrule
\end{tabular}
\end{table}

Table \ref{tab:fuzzysynth} demonstrates that the model essentially learns the three recommendation rules present in the synthetic dataset: \\
$HIGH \to RELEVANT$ (Rule 2 and 4), \\
$RECENT \land GENRE \to RELEVANT$ (Rule 1) and\\
$RECENT \land CAST \land DIRECTOR \to RELEVANT$ (Rule 3)\\
This is indicated by the high weights assigned to the respective atoms.
The low fuzzy weights of the remaining atoms suggest minimal impact on the learned rules.
In particular, the random atom COOKIE shows very low importance, indicating that only meaningful atoms are considered in the learned rules.
Moreover, this experiment demonstrates that if the selected number of rules $k$ surpasses the actual number of distinct rules in the dataset, some rules are learned redundantly.
This redundancy eliminates the need to know the true number of rules beforehand, as long as the number of stacked layers is larger than or equal to the actual number of rules.
The impact of removing these redundant rules on model performance will be explored in future work.

Since the learned rules match the ground truth rules, our model achieves values of $1.0$ for Precision@5, NDCG, and MAP (see Table~\ref{tab:results}).
In addition, we observe that the interpretable baselines JRip and Decision Tree can also learn the ground truth rules, leading to perfect evaluation results. The baseline NCF achieves the second-best results with close to perfect values, as shown in Table~\ref{tab:results}.

These results were achieved with $n=6$ predefined atoms, $k=4$ decision layers, a learning rate of $0.05$ over $300$ epochs, and a regularization parameter $\lambda=0.2$.

\begin{figure}[t]
  \centering
  \includegraphics[width=\linewidth,alt={
    This box plot illustrates the distribution of fuzzy weights for rules learned on the MovieLens 1M dataset. The box including the whiskers is below 0.1, indicating that most weights are very low. The three most important atoms give the outliers with high weights up to 0.7: HIGH AVG MOVIE RATING (brown), HIGH AVG RATING PER USER (cyan), and MOVIE RATED OFTEN (orange), showing their relative importance through color-coded points.
  }]{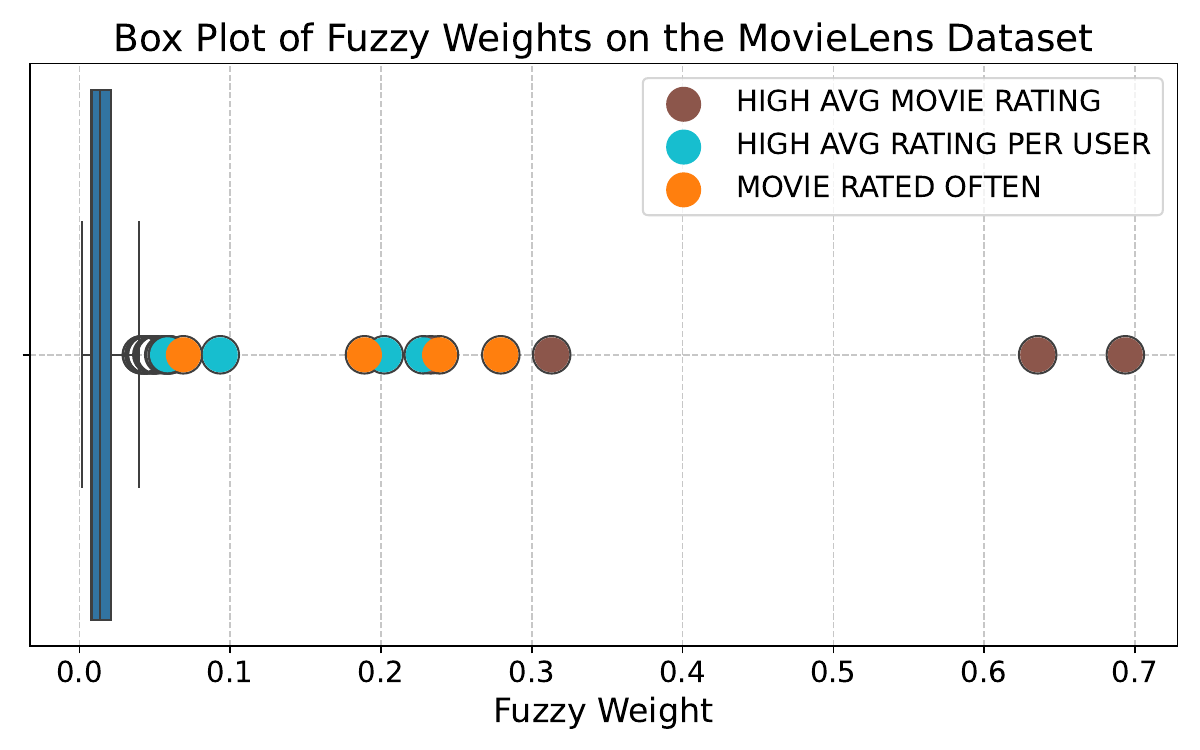}
  \caption{Box plot of the values in the fuzzy weight matrix $\mathbf{W'}$ on the MovieLens 1M dataset. We observe a clear difference in the weights of the three most important atoms - HIGH AVG MOVIE RATING, HIGH AVG RATING PER USER, and MOVIE RATED OFTEN - compared to the remaining 77 atoms.}
  \label{fig:boxplot}
\end{figure}

\subsection{MovieLens 1M Dataset}
\begin{table}[hb]
\caption{Truncated fuzzy weight matrix $\mathbf{W'}$ for MovieLens 1M for rules R1, R2, R3, R4. Values above 0.1 highlighted in bold.}
\label{tab:fuzzymovie}
\setlength{\tabcolsep}{3pt} 
\centering
\begin{tabular}{lcccc}
\toprule
 & $\,\ldots\,$ & \makecell{HIGH AVG MOVIE \\RATING (4.0+)} & \makecell{HIGH AVG RATING \\PER USER (4.0+)} & \makecell{MOVIE RATED \\OFTEN (228.0+)} \\
\midrule
R1 & $\,\ldots\,$ & $\mathbf{0.233}$ & $\mathbf{0.228}$ & $\mathbf{0.240}$ \\
R2 & $\,\ldots\,$ & $\mathbf{0.694}$ & $0.093$ & $0.068$ \\
R3 & $\,\ldots\,$ & $\mathbf{0.313}$ & $\mathbf{0.202}$ & $\mathbf{0.189}$ \\
R4 & $\,\ldots\,$ & $\mathbf{0.636}$ & $0.058$ & $\mathbf{0.279}$ \\
\bottomrule
\end{tabular}
\end{table}

\begin{table*}[t]
\caption{Evaluation results on the synthetic and MovieLens 1M datasets. The best-performing model for each metric is highlighted in bold, while the second-best model is underlined. Standard deviations for each metric are provided in parentheses alongside the values. The superscript * indicates results significantly different from our model, based on a two-sided Wilcoxon signed-rank test ($p<0.01$). The decision tree is evaluated twice — once using the original item and user features from the dataset (e.g., gender, age, movie genre) as input, and once using the predefined atoms as input.}
\label{tab:results}
\begin{tabular}{lllllllll}
\toprule
\multicolumn{9}{c}{Synthetic} \\
\midrule
 & P@5 & R@5 & NDCG@5 & MAP@5 & P@10 & R@10 & NDCG@10 & MAP@10 \\
\midrule
Our Model & $\mathbf{1.000_{(.000)}}$ & $\mathbf{.345_{(.000)}}$ & $\mathbf{1.000_{(.000)}}$ & $\mathbf{1.000_{(.000)}}$ & $\mathbf{.987_{(.000)}}$ & $\mathbf{.673_{(.000)}}$ & $\mathbf{1.000_{(.000)}}$ & $\mathbf{1.000_{(.000)}}$ \\
BaselineOnly & $.650_{(.000)}$* & $.213_{(.000)}$* & $.659_{(.000)}$* & $.571_{(.000)}$* & $.583_{(.000)}$* & $.379_{(.000)}$* & $.613_{(.000)}$* & $.473_{(.000)}$* \\
KNNBasic & $.475_{(.000)}$* & $.157_{(.000)}$* & $.476_{(.000)}$* & $.342_{(.000)}$* & $.474_{(.000)}$* & $.312_{(.000)}$* & $.477_{(.000)}$* & $.302_{(.000)}$* \\
SVD & $.734_{(.002)}$* & $.240_{(.001)}$* & $.769_{(.002)}$* & $.693_{(.002)}$* & $.603_{(.001)}$* & $.392_{(.001)}$* & $.670_{(.001)}$* & $.527_{(.002)}$* \\
LightGCN & $.472_{(.006)}$* & $.155_{(.002)}$* & $.472_{(.007)}$* & $.337_{(.007)}$* & $.472_{(.005)}$* & $.311_{(.003)}$* & $.475_{(.006)}$* & $.299_{(.005)}$* \\
JRip & $\mathbf{1.000_{(.000)}}$ & $\mathbf{.345_{(.000)}}$ & $\mathbf{1.000_{(.000)}}$ & $\mathbf{1.000_{(.000)}}$ & $\mathbf{.987_{(.000)}}$ & $\mathbf{.673_{(.000)}}$ & $\mathbf{1.000_{(.000)}}$ & $\mathbf{1.000_{(.000)}}$ \\
Decision Tree Orig & $\mathbf{1.000_{(.000)}}$ & $\mathbf{.345_{(.000)}}$ & $\mathbf{1.000_{(.000)}}$ & $\mathbf{1.000_{(.000)}}$ & $\mathbf{.987_{(.000)}}$ & $\mathbf{.673_{(.000)}}$ & $\mathbf{1.000_{(.000)}}$ & $\mathbf{1.000_{(.000)}}$ \\
Decision Tree Pred & $\mathbf{1.000_{(.000)}}$ & $\mathbf{.345_{(.000)}}$ & $\mathbf{1.000_{(.000)}}$ & $\mathbf{1.000_{(.000)}}$ & $\mathbf{.987_{(.000)}}$ & $\mathbf{.673_{(.000)}}$ & $\mathbf{1.000_{(.000)}}$ & $\mathbf{1.000_{(.000)}}$ \\
NCF & $\underline{.997_{(.000)}}$* & $\underline{.344_{(.000)}}$* & $\underline{.998_{(.000)}}$* & $\underline{.997_{(.001)}}$* & $\underline{.981_{(.001)}}$* & $\underline{.668_{(.001)}}$* & $\underline{.995_{(.001)}}$* & $\underline{.992_{(.001)}}$* \\
\midrule
\multicolumn{9}{c}{Movielens 1M} \\
\midrule
 & P@5 & R@5 & NDCG@5 & MAP@5 & P@10 & R@10 & NDCG@10 & MAP@10 \\
\midrule
Our Model & $.218_{(.001)}$ & $.225_{(.000)}$ & $.232_{(.001)}$ & $.210_{(.001)}$ & $.208_{(.000)}$ & $.343_{(.000)}$ & $.233_{(.000)}$ & $.204_{(.001)}$ \\
BaselineOnly & $\underline{.232_{(.000)}}$* & $\underline{.236_{(.000)}}$* & $\underline{.248_{(.000)}}$* & $\underline{.230_{(.000)}}$* & $\underline{.218_{(.000)}}$* & $\underline{.355_{(.000)}}$* & $\underline{.246_{(.000)}}$* & $\underline{.222_{(.000)}}$* \\
KNNBasic & $.202_{(.000)}$* & $.225_{(.000)}$ & $.217_{(.000)}$* & $.194_{(.000)}$* & $.191_{(.000)}$* & $.335_{(.000)}$* & $.218_{(.000)}$* & $.187_{(.000)}$* \\
SVD & $\mathbf{.235_{(.001)}}$* & $\mathbf{.238_{(.001)}}$* & $\mathbf{.251_{(.001)}}$* & $\mathbf{.234_{(.001)}}$* & $\mathbf{.220_{(.000)}}$* & $\mathbf{.357_{(.001)}}$* & $\mathbf{.249_{(.001)}}$* & $\mathbf{.225_{(.001)}}$* \\
LightGCN & $.173_{(.002)}$* & $.198_{(.002)}$* & $.185_{(.002)}$* & $.154_{(.002)}$* & $.167_{(.001)}$* & $.306_{(.001)}$* & $.189_{(.001)}$* & $.149_{(.001)}$* \\
JRip & $.192_{(.000)}$* & $.209_{(.000)}$* & $.204_{(.000)}$* & $.175_{(.000)}$* & $.185_{(.000)}$* & $.324_{(.000)}$* & $.207_{(.000)}$* & $.171_{(.000)}$* \\
Decision Tree Orig & $.188_{(.000)}$* & $.210_{(.000)}$* & $.199_{(.000)}$* & $.170_{(.000)}$* & $.179_{(.000)}$* & $.321_{(.000)}$* & $.202_{(.000)}$* & $.164_{(.000)}$* \\
Decision Tree Pred & $.179_{(.004)}$* & $.202_{(.002)}$* & $.191_{(.004)}$* & $.160_{(.004)}$* & $.175_{(.003)}$* & $.314_{(.003)}$* & $.196_{(.003)}$* & $.158_{(.004)}$* \\
NCF & $.208_{(.002)}$* & $.219_{(.001)}$* & $.222_{(.002)}$* & $.197_{(.003)}$* & $.195_{(.001)}$* & $.331_{(.001)}$* & $.221_{(.002)}$* & $.188_{(.002)}$* \\
\bottomrule
\end{tabular}
\end{table*}

Table~\ref{tab:fuzzymovie} shows the atoms with the highest fuzzy weights learned from the MovieLens 1M dataset; this paper does not depict atoms with weights $w'<0.1$ due to space constraints.
The box plot in Figure~\ref{fig:boxplot} illustrates the distribution of values in the fuzzy weight matrix $\mathbf{W'}$, clearly indicating that the majority of fuzzy weights is below $0.1$.
Additionally, we can observe that the atom HIGH AVG MOVIE RATING is the most important across the rules, followed by MOVIE RATED OFTEN and HIGH AVG RATING PER USER.
The impact of the atom MOVIE RATED OFTEN suggests that the data is biased towards popular movies.
Moreover, we find that 77 atoms, such as those related to gender, occupation, age, and movie genres, have weights $w'<0.1$ and thus contribute only weakly to the learned rules. 
An example of a rule learned by our model recommends popular and highly rated movies, derived from rule 4 (see Table~\ref{tab:fuzzymovie}). We present a simplified form of the learned fuzzy rule by omitting atoms with weights $w'<0.1$. This is only a post-hoc interpretation step and does not affect the model's predictions.
\begin{align*}  
\text{HIGH AVG MOVIE RATING } \land \text{ MOVIE RATED OFTEN } \\
\to RELEVANT
\end{align*}
We can observe that the fuzzy weights derived from the MovieLens dataset (see Table~\ref{tab:fuzzymovie}) are notably lower compared to the weights of the synthetic dataset (see Table~\ref{tab:fuzzysynth}). We attribute this to atoms with $w'<0.1$ still influencing the decision rule, but at the same time, potentially reducing the relative importance of atoms with $w'\geq0.1$. 

To determine appropriate thresholds for the atoms HIGH AVG MOVIE RATING, HIGH AVG RATING PER USER, and MOVIE RATED OFTEN, the 10th, 25th, 50th, 75th, and 90th percentiles of these values (based on the training data) were considered.
For each threshold, an atom with this threshold was added to the predefined atoms.
The atom with the highest importance, determined by the highest fuzzy weights, was then retained while those with lower thresholds were removed.
To handle cold-start users and items with no rating history, we assign their corresponding atom values to dataset means, i.e., effectively assuming a user and item bias of $0$.

The learned rules align with the evaluation results in Table \ref{tab:results}, where the BaselineOnly approach, which combines mean rating with user and item bias, achieves the second-best performance after the matrix-factorization model SVD.
In addition, our learned fuzzy neural network achieves significantly better results than the interpretable baselines JRip and Decision Tree and the NCF model on all metrics, which indicates that the learned rules can explain the user behavior well. 
Nevertheless, SVD significantly outperforms our model and all other baselines across all metrics.


The evaluation results of our approach on the MovieLens 1M dataset were achieved with $n=80$ predefined atoms, $k=4$ decision layers, a learning rate of $0.05$ over $150$ epochs, and a regularization parameter $\lambda=0.1$.

\section{Conclusion}
In this work, we present a fuzzy neural network approach for transparent recommendations that combines symbolic and sub-symbolic mechanisms.
Our model learns logic rules over predefined, human-readable atoms using fuzzy logic operators, enabling transparent reasoning paths that mimic human-like decision-making. On a synthetic dataset, our model accurately learns ground-truth rules. On the MovieLens 1M dataset, it achieves competitive performance compared to standard baseline recommendation algorithms.
While our approach does not outperform matrix factorization in top-k metrics, it prioritizes transparency, a critical property in human-centric recommender systems. Overall, our model represents a promising step toward transparent recommender systems that are effective and inherently understandable.  

\paragraph{\textbf{Limitations.}} We acknowledge several limitations in our current work.  First, while our model is transparent by design, we have not yet conducted user studies to evaluate how effectively end users can understand and interact with the learned rules. Such studies are essential to validate the interpretability and usability of the system in practice.
Second, our evaluation has so far been limited to a synthetic dataset and the MovieLens 1M dataset. While these provide useful insights, further experiments on more real-world datasets are required to assess generalizability, robustness, and performance across domains.

Third, the quality and clarity of the learned rules depend on the selection of predefined atoms and the behavior of the training process. Since rule learning is influenced by the random initialization of weights, different training runs may yield different sets of rules, potentially affecting consistency. Currently, there is no mechanism for selecting or verifying learned rules, which would be valuable for ensuring reliability.
Fourth, our model relies on predefined atoms, which may be incomplete for users or items with limited metadata or interaction history. Cold-start performance is thus limited.
Finally, our model is not explicitly optimized for ranking. While we use fuzzy outputs as confidence scores to produce ranked item lists, incorporating ranking-aware objectives such as pairwise or listwise ranking loss functions could improve top-k performance without compromising transparency.

\paragraph{\textbf{Future Work.}} In our experiments, we find that a large fraction of learned fuzzy weights are below $0.1$, indicating low relative importance for many atoms. However, a systematic evaluation of pruning strategies and thresholds and their effects remains future work.
To improve evaluation and assess the generalizability of our approach, we plan to evaluate the model's performance on more datasets. As our model has only been tested on a moderate number of atoms and rules, we plan to evaluate the scalability of this approach on more complex datasets. We also aim to conduct a comprehensive sensitivity analysis to explore the effects of hyperparameters, dataset characteristics - such as rule support - and the choice of t-norms on performance and transparency. 
In addition, we plan to explore how to adapt the learned rules to individual users, e.g., by fine-tuning rules on user interaction histories.
We also aim to investigate the effect of removing atoms related to item popularity on fairness. 
To reduce the need for predefined atoms, we plan to integrate interpretable encoder neural networks or decision trees for learning additional atoms directly from user-item interactions and metadata. This could enhance the capacity of the model to capture more complex behavioral patterns.
Finally, we plan to conduct user studies to assess how users interpret and engage with the learned rules. Our long-term goal is to use these rules to inform the design of user-centered explanations. 

\begin{acks}
This research is funded by: Austrian Science Fund 10.55776/COE12, Cluster of Excellence {\textcolor{blue}{\href{https://www.bilateral-ai.net/home}{Bilateral Artificial Intelligence}}} and FFG HybridAir project (project \#FO999902654).
 
\end{acks}

\bibliographystyle{ACM-Reference-Format}
\bibliography{main}

\end{document}